\documentclass[10pt,twocolumn,letterpaper]{article}

\usepackage{iccv}              % To produce the CAMERA-READY version
\usepackage{url}            % simple URL typesetting
\usepackage{booktabs}       % professional-quality tables
\usepackage{amsfonts}       % blackboard math symbols
\usepackage{nicefrac}       % compact symbols for 1/2, etc.
\usepackage{microtype}      % microtypography
\usepackage{array}
\usepackage{gensymb}
\usepackage{amsmath}
\usepackage{amsthm}
\usepackage{amssymb}
\usepackage{graphicx}
\usepackage[accsupp]{axessibility}
\usepackage{multirow}
\usepackage{makecell}
\usepackage{caption}

\usepackage{adjustbox}
\usepackage{wrapfig}

\definecolor{iccvblue}{rgb}{0.21,0.49,0.74}
\usepackage[pagebackref,breaklinks,colorlinks,allcolors=iccvblue]{hyperref}

\def\paperID{5413} % *** Enter the Paper ID here
\def\confName{ICCV}
\def\confYear{2025}

\title{VideoAds for Fast-Paced Video Understanding}

\author{
Zheyuan Zhang$^{1}$\footnotemark[1] \quad 
Wanying Dou$^{1}$\footnotemark[1] \quad 
Linkai Peng$^{1}$\footnotemark[1] \quad 
Hongyi Pan$^{1}$ \quad 
Ulas Bagci$^{1}$ \quad 
Boqing Gong$^{2}$ \\ 
\small $^{1}$Northwestern University \quad $^{2}$Boston University \\
\tt\small \{zheyuan.zhang, monica.dou, linkai.peng, hongyi.pan, ulas.bagci\}@northwestern.edu, 
\tt\small bgong@bu.edu
}
\begin{document}
\maketitle

\begin{abstract}

Advertisement videos serve as a rich and valuable source of purpose-driven information, encompassing high-quality visual, textual, and contextual cues designed to engage viewers. They are often more complex than general videos of similar duration due to their structured narratives and rapid scene transitions, posing significant challenges to multi-modal large language models (MLLMs). In this work, we introduce VideoAds, the first dataset tailored for benchmarking the performance of MLLMs on advertisement videos. VideoAds comprises well-curated advertisement videos with complex temporal structures, accompanied by \textbf{manually} annotated diverse questions across three core tasks: visual finding, video summary, and visual reasoning. We propose a quantitative measure to compare VideoAds against existing benchmarks in terms of video complexity. Through extensive experiments, we find that Qwen2.5-VL-72B, an open‑source MLLM, achieves 73.35\% accuracy on VideoAds, outperforms GPT-4o (66.82\%) and Gemini-1.5 Pro (69.66\%); the two proprietary models especially fall behind the open‑source model in video summarization and reasoning, but perform the best in visual finding. Gemini-2.5 Pro leads with an accuracy of 80.04\%. Notably, human experts easily achieve a remarkable accuracy of 94.27\%. These results underscore the necessity of advancing MLLMs' temporal modeling capabilities and highlight VideoAds as a potentially pivotal benchmark for future research in understanding video that requires high FPS sampling. The dataset and evaluation code will be publicly available at \url{https://videoadsbenchmark.netlify.app}.

\end{abstract}    
\section{Introduction}
\label{sec:intro}
\begin{figure}[h]
    \centering
    \includegraphics[width=0.8\linewidth]{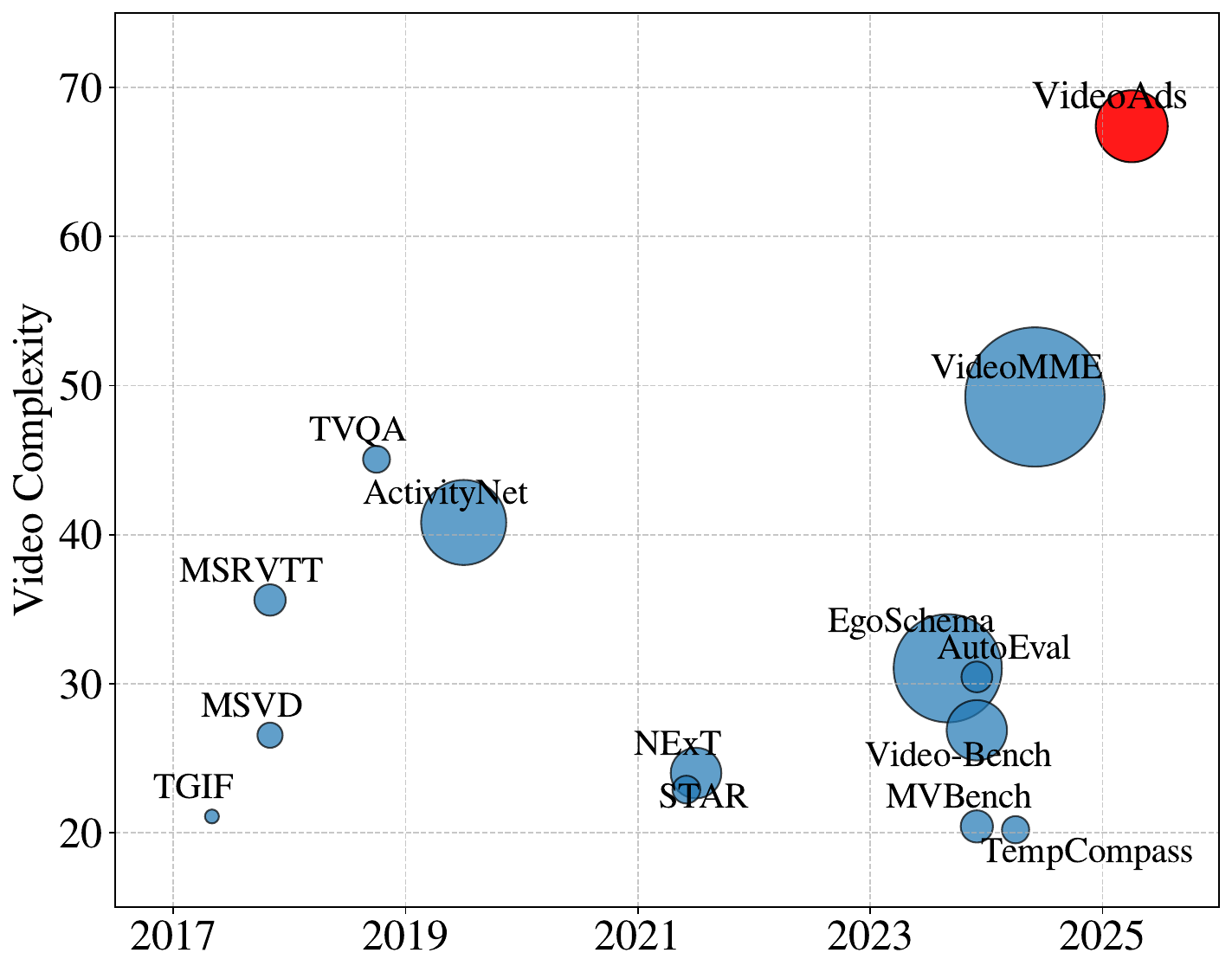}
    \caption{Recent years have witnessed a significant increase in the complexity of video benchmarks, paralleling the rapid progress in the capabilities of multi-modal large language models (MLLMs). In this work, we introduce VideoAds, a complex dataset based on advertisement videos, specifically designed to benchmark the performance of MLLMs on challenging visual comprehension and complex temporal reasoning. The size of each scatter point represents the average duration of videos within each dataset.}
    \label{fig:temporal_evolution}
    \vspace{-4mm}
\end{figure}

\begin{figure*}[t]
    \centering
    \includegraphics[width=0.8\linewidth]{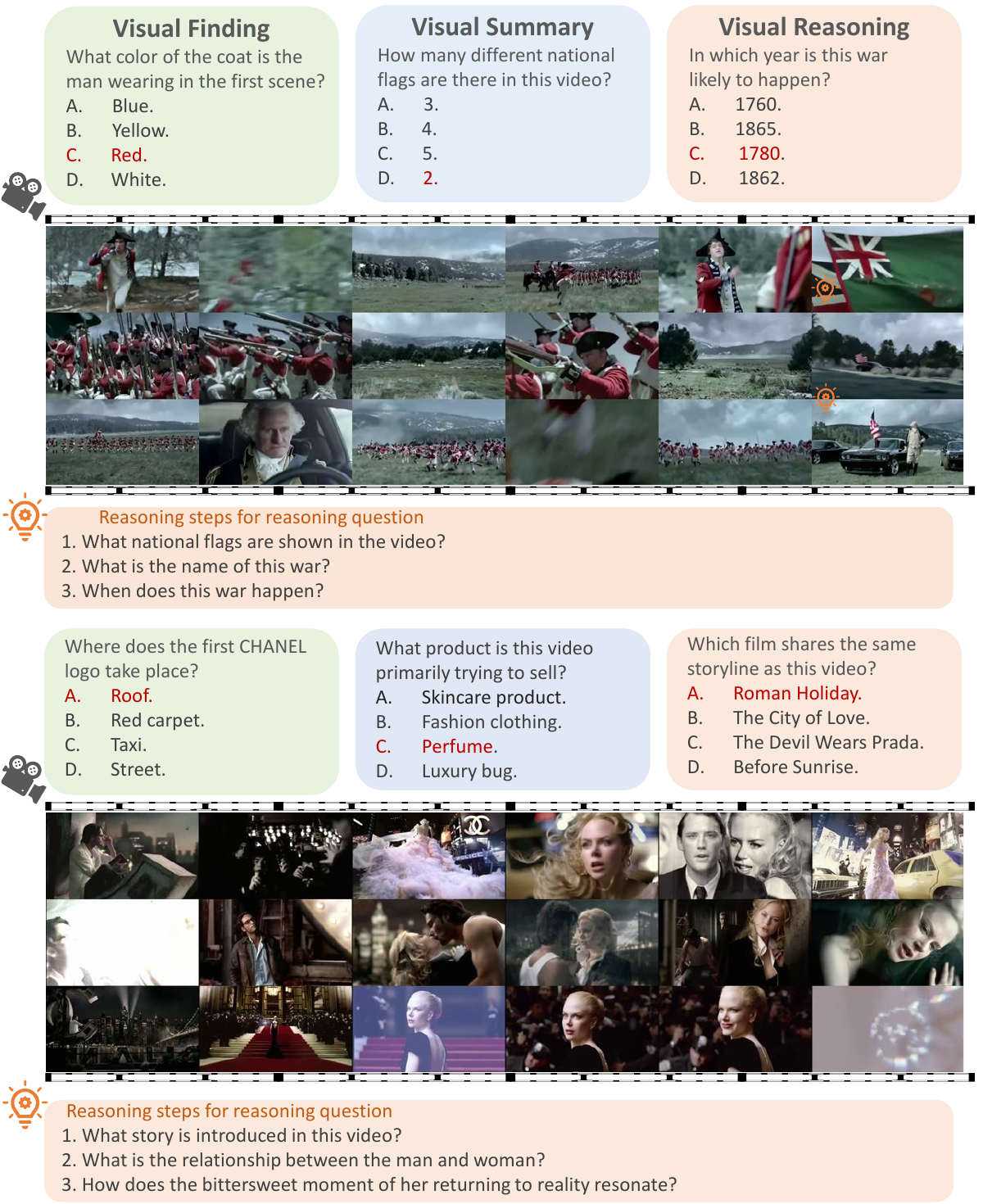}
    \caption{VideoAds comprises three challenging tasks: Visual Finding, Visual Summary, and Visual Reasoning, specifically designed to evaluate MLLMs' temporal reasoning capabilities on videos with complex temporal structures that have never been investigated before. Unlike many previous datasets that focus on recognizing isolated actions or events, VideoAds demands that models derive the correct answers only through multistep reasoning over multi-modal visual clues. 
    }
    \label{fig:visual_dataset}
    \vspace{-4mm}
\end{figure*}

Advertisement videos represent a unique and high-value segment of visual media, carefully crafted to capture the audience's attention through rich multimodal content in a short time~\cite{yang2017consumerads}. Unlike general videos, advertisements integrate carefully designed storytelling, persuasive visual elements, and concise yet information-dense narratives to convey targeted messages efficiently. These videos often include high-quality cinematography, strategic scene transitions, and contextual cues that enhance engagement and retention, making them invaluable in marketing, entertainment, and e-commerce. %Consequently, the ability to understand, summarize, and reason over advertisement videos is not merely beneficial but crucial for real-world applications, including automated content analysis, targeted ad retrieval, and predictive consumer behavior modeling. %\boqing{These real-world applications are good, but I feel we need even stronger argument about the importance of understanding ads video}. 

The rapid advancements in Large Language Models (LLMs)~\cite{gpt3,reid2024gemini} have revolutionized multi-modal video understanding, enabling models to process and reason over video content with unprecedented efficacy~\cite{tang2023video,zhang2024mm,hurst2024gpt4o}. Generally, these models take videos as input and project them into the embedding or token space of LLMs~\cite{zhang2024llavanextvideo,openai2023gpt4v,reid2024gemini,Ormazabal2024RekaCF} and leverage LLMs to generate high-quality outputs. These multi-modality LLMs (MLLMs) exhibit remarkable open-ended generation capabilities, including but not limited to temporal, spatial, and causal inference, making them promising for real-world video applications~\cite{fu2024videomme,li2023mvbench}.

However, we contend that existing video-language benchmarks do not thoroughly assess MLLMs' ability to comprehend complex temporal dependencies. Advertisement videos, a valuable but underexplored video type in MLLM evaluation, while typically short, exhibit intricate multi-stage narratives, rapid scene transitions, and implicit causal relationships, making them particularly difficult for MLLMs to comprehend. In contrast, existing benchmarks for video understanding primarily focus on generic action recognition, instructional videos, movie comprehension, and general videos collected from YouTube~\cite{fu2024videomme,Liu2024TempCompassDV,mangalam2024egoschema,li2023mvbench}. These videos are rich and diverse in actions and scenes, but they are not comparable to advertisement videos in terms of how to purposely use complex visual cues, background context, and temporally disjointed sequences to convey persuasive messaging. As a result, advertisement videos are a more challenging testbed than generic videos for assessing MLLMs' inference ability across time and space, multi-step reasoning about narratives, and long-range summarization, among other essential abilities needed for video understanding. % Thus, despite their success in other video types, most existing MLLM models fail to capture advertisements' multi-modal signals embedded in information-rich video sequences with complex temporal structures.

% This gap in evaluation frameworks hinders the development of models capable of comprehensively interpreting multi-modal signals embedded in information-rich video sequences with complex temporal structures. %\boqing{It's a little dangerous to charactorize ads video into short-form, information-rich videos because many TikTok videos have those properties, too}
To this end, we introduce VideoAds, the first dedicated dataset for evaluating MLLMs in the context of advertisement video understanding. It is important to position VideoAds in the landscape of existing benchmarks. VideoAds is not intended to replace larger, more comprehensive benchmarks like Video-MME~\cite{fu2024videomme}, but rather to complement them. VideoAds presents unique challenges for video understanding due to its complex temporal structures and high-density semantic content, making it a testbed with high potential for identifying limitations of current MLLMs in temporal reasoning and, accordingly, driving the development of next-generation AI systems with advanced temporal modeling, cross-modal interactions, and multi-stage reasoning capabilities. Beyond academic impact, VideoAds addresses real-world challenges in automated content analysis, brand sentiment prediction, and strategic ad placement, facilitating AI-empowered applications in the advertisement sector. %paving the way for AI systems capable of comprehending complex narratives and implicit causal relationships. By setting a new standard for complex video understanding, VideoAds bridges the gap between academic research and industry needs, fostering innovation and progress in multi-modal AI systems.%\boqing{Probably expand the introduction a little bit by including the potential impacts of our work. Otherwise, a reviewer could say: okay, a new benchmark, so what? What future could it drive the community to?} 

\noindent We summarize our main contributions as follows:
\begin{itemize}
    \item \textbf{First Advertisement Video Benchmark for MLLMs}: We present VideoAds, a carefully curated benchmark dataset consisting of high-quality and high-complexity advertisement videos, featuring \textbf{manually} annotated questions covering three groups of tasks: visual finding, video summarization, and visual reasoning.
    \item \textbf{Quantitative Video Complexity Measure}: We introduce a novel quantitative measure for video complexity, enabling a holistic analysis of how scene transitions, narrative coherence, and temporal event dependencies influence MLLM performance. This measure also accommodates a comparison between VideoAds and existing video understanding datasets in terms of complexity.
    \item \textbf{Comprehensive Benchmarking of State-of-the-Art Models}: We evaluate leading MLLMs, including GPT-4o~\cite{openai2024gpt4o}, Gemini 1.5 Pro~\cite{team2023gemini}, and open source models like LLaVa-Video~\cite{zhang2024llavanextvideo}, Qwen2.5-VL~\cite{Qwen2VL} over multiple-choice VQA tasks, revealing significant gap from human performance when the models attempt to reason across complex temporal structures. Moreover, MLLMs perform well in finding visual patterns from videos but struggle with video summarization and reasoning.%, while the opposite is true for humans.
    \item \textbf{Insights about Audio and Chain of Thought in MLLMs}: We conduct further analyses about the influence of speech audio transcripts and Chain of Thoughts (CoT). We find that large MLLMs benefit from CoT more than small models, aligning with LMMs' properties, and the additional information from speech transcripts almost always improves MLLMs' performance. These results signify the need for future MLLMs to reinforce cross-modal interactions and reasoning.
\end{itemize}

%-------------------------------------------------------------------------
\noindent

\begin{figure*}[h]
    \centering
    \includegraphics[width=0.8\textwidth]{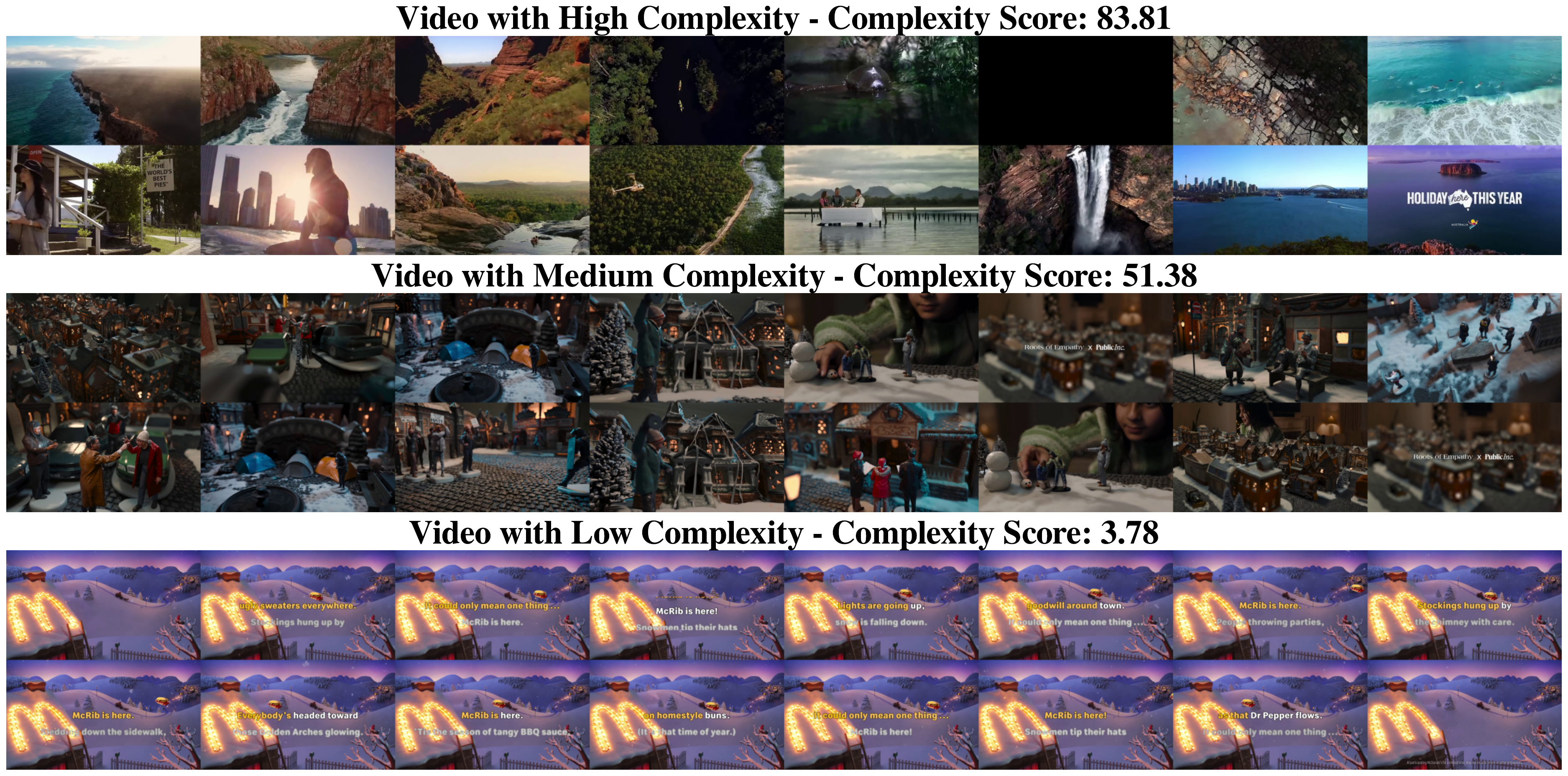}
    \caption{Visualization of Video Complexity Scores: Examples of videos with high, medium, and low complexity scores based on the proposed video complexity metrics. Videos with high complexity scores exhibit frequent scene transitions, dynamic interactions, and complex visual transformations, posing significant challenges for temporal reasoning. In contrast, videos with low complexity display static or minimally changing frames, resembling image slideshows with minimal narrative progression.
    }
    \label{fig:complexity_visual}
    %\vspace{-2mm}
\end{figure*}

\section{What makes an advertisement video?}
Advertisement videos are a unique form of visual storytelling, meticulously crafted to capture audience attention, convey persuasive messages, and influence consumer behavior within a short time~\cite{ikonomi2019tv,smethurst2009write}. %Unlike traditional video content, advertisements are strategically designed to create brand awareness and drive sales. Their distinctive narrative structure and high-density information make them exceptionally valuable. Consequently, advertisement videos are significantly more expensive than other video types due to high production quality, professional cast, and strategic creative direction. 
For example, the second video in Figure~\ref{fig:visual_dataset}, ``CHANEL N\degree5 the Film'' is a 120-second short advertisement directed by Baz Luhrmann and with a budget of  \$33 million~\cite{enwiki:1266185410}. This advertisement tells a story: ``A famous celebrity runs away in a pink dress in the middle of Times Square in New York City, only to get into a cab with a man who does not recognize her. After four days in his Lower East Side apartment, her secretary insists that she return as a celebrity.'' This advertisement video shares a plotline similar to Roman Holiday but condenses it into just 120 seconds and provides us with high-density information that has rarely been seen in other video types. %Similarly, most advertisements follow one storyline to introduce the product in 30 to 90 seconds due to the display cost requirements. 
The high commercial value translates into complex narrative structures and information-dense sequences, making advertisement videos more challenging to analyze for the current MLLMs which often rely on low FPS sampling.

\section{VideoAds}
\label{sec:videoAds}
To build the VideoAds benchmark, we take three steps: video collection followed by manual filtering; curation of VQA tasks; and quality review.
\begin{table*}[!ht]
\centering
\caption{The comparison of various minutes long video benchmarks: total number of videos (\textbf{\#Videos}), number of clips (\textbf{\#Clips}), average duration of the videos (\textbf{Len.}), number of QA pairs (\textbf{\#QA Pairs}), method of annotation (\textbf{Anno.}, M/A means the manually/automatic manner), average number of QA pair tokens (\textbf{QA Tokens}), complexity duration(\textbf{Complexity Duration}), video complexity (\textbf{Video Complexity})} 
\resizebox{0.8\textwidth}{!}{
\begin{tabular}{lrrrrc|cc}
\toprule
\textbf{Benchmarks} & \textbf{\#Videos} & \textbf{\#Clips} &\textbf{Len.(s)} & \textbf{\#QA Pairs} &  \textbf{Anno.} & \textbf{Complexity Duration ($\uparrow$)} & \textbf{Video Complexity ($\uparrow$)} \\
\midrule
TGIF-QA~\citep{jang2017tgif} & 9,575 & 9,575 & 3.0 & 8,506 & A\&M & 0.67 & 21.10 \\
STAR~\citep{wu2021star_situated_reasoning} & 914 & 7,098 & 11.9 & 7,098 & A & 6.25 & 22.90 \\
NExT-QA~\citep{xiao2021next} & 1,000 & 1,000 & 39.5 & 8,564 & A & 9.23 & 24.01 \\
MSVD-QA~\citep{Xu2017VideoQA}& 504 & 504 & 9.8 & 13,157 & A  & 1.75 & 26.55 \\
MSRVTT-QA~\citep{Xu2017VideoQA} & 2,990 & 2,990 & 15.2 & 72,821 & A & 3.68 & 35.62 \\
ActivityNet-QA \citep{yu2019activitynet} & 800 & 800 & 111.4 & 8,000 & M & 47.59 & 40.82 \\
% TVQA~\citep{lei2018tvqa} & 2,179 & 15,253 & 11.2 & 15,253 & M & 26.19 & 45.06\\
\midrule
TempCompass~\citep{Liu2024TempCompassDV} & 410 & 500 & 11.4 & 7,540 & A\&M & 1.62 & 20.21\\
MVBench~\citep{li2023mvbench} & 3,641 & 3,641 & 16.0 & 4,000 & A & 3.34 &  20.43 \\
Video-Bench~\citep{ning2023video} & 5,917 & 5,917 & 56.0 & 17,036 & A\&M & 16.58 & 26.88 \\
EgoSchema~\citep{mangalam2024egoschema} & 5,063 & 5,063 & 180.0 & 5,063 & A\&M & 54.30 & 31.04 \\
AutoEval-Video~\citep{chen2023autoeval} & 327 & 327 & 14.6 & 327 & M  & 3.46 & 30.46 \\
Video-MME-S~\cite{fu2024videomme} & 300 & 300 & 80.7 & 900 &  M & 44.81 & 56.10  \\
Video-MME-M~\cite{fu2024videomme} & 300 & 300 & 515.9 & 900 & M & 220.88 & 42.38\\
% Video-MME S\&M~\cite{fu2024videomme} & 600 & 600 & 298.3 & 1800 & M & 49.80 \\
\midrule
VideoAds & 200 & 200 & 79.6 & 1100 & M  & 60.32 &  67.40 \\
\bottomrule
\end{tabular}}
\vspace{-4mm}
\label{tab:comparison}
\end{table*}

\subsection{Video collection and filtering}
We begin by collecting YouTube advertisement videos, using targeted keyword searches to encompass various commercial domains. Specifically, we retrieve videos using commercial search terms about Cars, Sports, Movie Trailers, Food, Tech, Health, Travel, and Financial Services. This broad list ensures our dataset covers multiple industries, allowing for diverse visual storytelling techniques, product promotions, and narrative-driven ad structures. After this step, we collected 2,700 videos from YouTube. 

To refine the dataset, we implement a systematic filtering process to remove irrelevant, excessively short, low-quality, and static-content videos that do not align with our benchmarking objectives. The specific filtering criteria are attached in the Appendix. After applying these filtering steps, we keep 200 high-quality, visually complex advertisement videos from the original 2,700 candidates. The data size of 200 is a sweet trade-off between the benchmark's coverage (in terms of diversity and visual quality) and utility (in computational and financial costs).
These videos have densely packed content, including but not limited to rich visual storytelling, diverse editing styles, and complex temporal structures, making them a versatile testbed for evaluating the temporal reasoning capabilities of MLLMs. %The resulting benchmark dataset is challenging ability to understand multistage event progressions, implicit causal relationships, and cross-modal between audio and vision interactions within short but densely packed video content.

\subsection{VQA annotations}
We ask human experts to manually construct video question answering (VQA) tasks about the videos. The annotation process is designed to emphasize complex temporal reasoning, visual interpretation, and high-level narrative understanding, addressing key aspects of video understanding. We categorize the VQA tasks into three core types.
\begin{itemize}
    \item \textbf{Visual finding}: This category assesses the model’s ability to extract concrete visual patterns from video frames.
    It includes but is not limited to object recognition, attribute identification (\eg, color, shape, size), detecting spatial relationships among objects, and scene recognition. Some exemplar questions are as follows. ``What color is the second car in the advertisement?''  ``Where is the ball on the table?'' ``In which city does the first scene take place?''

    \item \textbf{Visual summary}: This category evaluates the model’s ability to summarize key events and thematic elements in the video. Questions focus on event description, sequential understanding, changes across time, and topic discovery. For example, ``How many types of insurance are mentioned in the ads?'', ``What is the main product being advertised?'', ``What happened before the targeted product is introduced?"

    \item \textbf{Visual reasoning}: This is the most challenging category, requiring the model to perform high-order  inference based on the video’s context. The questions involve cause-and-effect analyses, interpersonal dynamics, emotional interpretation, and logical problem-solving. For example, ``Why does the man finally decide to purchase the product?'', ``What emotion does the actor express after using the product?'', ``How does the advertisement convey the product’s benefits using different scenes?''
    \vspace{5pt}
\end{itemize}

While semi-automated VQA annotation methods using powerful MLLMs~\cite{yan2021videogpt,maaz2024videogpts} can easily generate a large number of question-answer pairs using human-provided captions, they are inherently limited by the granularity of the captions and MLLMs' potential bias. %, evidently favoring static visual patterns over complex temporal reasoning. %Pretrained VLMs excel at recognizing objects, attributes, and scene descriptions but struggle with interpreting multi-stage narratives where the answer depends on long-range event dependencies and understanding implicit causal relationships. For example, a VLM may accurately describe the color of a car in a commercial but fail to infer why the driver’s behavior changes after interacting with the product.
Hence, in this work, the VQA tasks are mainly generated by human annotations. Human annotators are required to create a question, a correct answer, and an incorrect (yet confusing) answer for each VQA task. For every video, an annotator is required to generate 10 VQA pairs, leading to 2,000 VQA tasks in total. We then employ the LLava-Next-72B model~\cite{zhang2024llavanextvideo} to generate four candidate incorrect answers. Then, OpenAI GPT4o~\cite{hurst2024gpt4o} is used to merge the question, correct answer, and human-provided and model-generated incorrect answers into standardized four-option multiple-choice questions. We show examples of all stages of this annotation process in the Appendix.

\subsection{Quality review}

To ensure the accuracy and depth of the generated VQA tasks, we apply a quality control step to preserve the most informative and challenging questions, ensuring a balanced distribution across the three primary VQA categories and removing trivial queries. Specifically, we remove VQA pairs that are trivial or overly simplistic, as well as those that contain ambiguities, multiple correct answers, or annotation errors. After expert filtering, we retain 1,100 high-quality VQA tasks from the original 2,000. The numbers of visual finding, visual summary, and visual reasoning tasks are 425, 312 and 363, respectively, and evenly distributed across all question types. Overall, this refinement significantly improved the VQA quality, making it a challenging benchmark for assessing MLLMs. % By removing trivial queries and emphasizing complex temporal structures, the quality review ensures that VideoAds effectively highlights the fundamental limitations of current VLMs in understanding advertisement videos.

\subsection{Dataset statistics}
We analyze VideoAds and contrast it with existing ones and, through this comparison, we notice that VideoAds differs from others primarily by its high visual complexity across time. The following first defines video complexity, a quantitative measure, and then presents dataset statistics.

\subsubsection{Definition of video complexity}

Although many video datasets claim to collect complex videos from diverse application scenarios, a quantitative definition of video complexity remains elusive in video analysis. %Recent studies, such as~\cite{tong2024eyes} and~\cite{Liu2024TempCompassDV}, reveal that multi-modal large language models (MLLMs) struggle to capture temporal changes accurately, highlighting a fundamental limitation in current video understanding methodologies. 
Inspired by~\cite{tong2024eyes,Liu2024TempCompassDV} and moving beyond qualitative observations, we propose a quantitative measure of video complexity based on the variance of visual features over time, addressing the need for a structured approach to measuring complexity in MLLM-based video understanding.

Intuitively, a video that exhibits greater changes over time within a given duration is more complex and challenging to understand than a static or minimally changing video. In the context of MLLMs, this translates to greater difficulty in understanding temporal coherence/changes, causal reasoning, and narrative understanding. %Therefore, by quantifying the degree of visual variance, we can more accurately evaluate a model’s ability to process complex temporal structures, thus paving the way for advancements in video-based AI reasoning. 
In this work, we define video complexity based on visual feature variance with DINO-v2 (referred to as $f$) as a feature extractor~\cite{oquab2023dinov2}. The reason for choosing DINO rather than CLIP~\cite{radford2021clip} is that DINO preserves more visual details than CLIP~\cite{tong2024eyes}. %We further provide some failure cases using CLIP in Appendix. 

Given a specific video $V$, we sample one frame $I_i$ per second from it and denote by $n$ the total number of video frames. Denote by $[-d, d]$ the neighbor of any frame. We define \emph{complexity duration} $D_{cpx}$ as:
\begin{equation}
    % D_{cpx} = \sum_i^{n}\frac{\sum_{j=\max(i-d, 1)}^{\min(i+d, n)} e^{-\frac{|j-i|}{2d}} (1-\cos(f(I_i), f(I_j)))}{\sum_{j=\max(i-d, 1)}^{\min(i+d, n)} e^{-\frac{|j-i|}{2d}}},
    D_{cpx} = n - \sum_{i=1}^{n}\frac{\sum_{j=\max(i-d, 1)}^{\min(i+d, n)} e^{-\frac{|j-i|}{2d}} \cos(f(I_i), f(I_j))}{\sum_{j=\max(i-d, 1)}^{\min(i+d, n)} e^{-\frac{|j-i|}{2d}}},
    \label{eqa:tc_define}
\end{equation}
where $j \neq i$, $\cos(., .)$ indicates cosine similarity, and the exponential weighting $e^{-\frac{|j-i|}{2d}}$ ensures that nearby frames contribute more to the complexity calculation than distant frames. The complexity duration is named in analogy to time duration. Indeed, Egoschema~\citep{mangalam2024egoschema} videos are long in duration (180 seconds per video, on average), but their average complexity duration is 54.3. We further define \emph{video complexity} $V_{cpx}$ (also called complexity density) as the average of complexity duration across time:
\begin{equation}
    V_{cpx} = {100}D_{cpx}/n,
    \label{eqa:vc_define}
\end{equation}
A high video complexity $V_{cpx}$ value indicates great visual changes over unit time, necessitating high FPS sampling strategies for MLLMs to comprehend the video thoroughly. Figure~\ref{fig:complexity_visual} illustrates videos with high, medium, and low video complexities, validating our measure. %Videos with high complexity scores contain frequent scene transitions, dynamic interactions, and complex visual transformations, demanding advanced temporal reasoning. Conversely, videos with low complexity scores appear more static, resembling image slideshows rather than dynamic video sequences. 
%This visualization further validates the effectiveness of the proposed metric in measuring temporal complexity in real-world scenarios. Additionally, 
We analyze the impact of neighborhood size $d$  in the Appendix.

\subsubsection{VideoAds analysis and comparison with others}

VideoAds distinguishes itself from existing video benchmarks by presenting a unique combination of high video complexity and challenging question-answer (QA) pairs, despite containing relatively short videos: videos in \cite{fu2024mmeevalsurvey} are up to hours long. Unlike conventional datasets, such as MSVD-QA~\cite{Xu2017VideoQA}, Youcook2~\cite{zhou2018youcook}, and ActivityNet-QA~\cite{yu2019activitynet}, which primarily focus on action recognition or object identification with lower video complexity, VideoAds emphasizes complex temporal reasoning on challenging tasks. This is achieved by curating advertisement videos that compress rich visual narratives and multi-stage events into one to two minutes, demanding long-range contextual tracking and multi-hop reasoning. Compared to Video-MME~\cite{fu2024videomme}, MMBench-Video~\cite{liu2023mmbench}, and recent Video-MMMU~\cite{hu2025videommmu}, which focus on longer video durations or multi-domain understanding, VideoAds presents a higher density of temporal changes. Table~\ref{tab:comparison} uses the video complexity measure to highlight that VideoAds videos contain significantly more temporal dynamics than other benchmarks. This compact complexity makes it particularly challenging for MLLMs to maintain temporal coherence and causal reasoning, even more so than processing long-form but slowly progressing videos. Furthermore, unlike datasets~\cite{liu2023mmbench,fu2024mmeevalsurvey} that use free-form questions, VideoAds employs four-option multiple-choice questions facilitating easy evaluations across different models. Finally, Table~\ref{tab:basic_statics} presents some statistics for each type of question. In summary, VideoAds raises the bar in video understanding by introducing complex and diverse advertisement videos and VQA tasks.

It is crucial to understand why benchmarks like Temp-Compass and MVBench  score lower on our complexity metric despite being used for fine-grained temporal understanding. Our metric, based on DINO-v2 feature variance, excels at capturing coarse visual shifts and significant content changes typical of advertisements with rapid cuts between distinct scenes. In contrast, videos in TempCompass or MVBench often feature more subtle temporal changes (e.g., an object moving slightly), which results in simpler overall video content and thus a lower complexity score. VideoAds therefore, serves as a complementary benchmark, focused on challenging models with fast-paced, semantics-rich visual transitions rather than fine-grained temporal shifts.

\begin{table}
    \centering
    \resizebox{0.8\linewidth}{!}{
    \begin{tabular}{l|ccc|c}
    \toprule
      & Finding & Summary & Reasoning & Total \\
    \midrule 
    \#Question  & 9.58 & 8.85 & 9.32 & 9.29 \\
    \#Options  & 33.64 & 40.38 & 50.09 & 40.98 \\
    Duration & 83.28 & 78.45 & 78.01 & 79.60 \\
    QA Tokens  & 53.63 & 59.95 & 71.80 & 61.42 \\
    \midrule
    Number  & 425 & 312 & 363 & 1100\\
    \bottomrule
    \end{tabular}}
    
    \caption{Summary of key statistics across the three question types in VideoAds, including average word count for questions (\#Question), answer options (\#Options), as well as average video duration (Duration) and average QA tokens (QA Tokens).}
    \label{tab:basic_statics}
    \vspace{-6mm}
\end{table}

\section{Experiments and Results}
\label{sec:results}

\begin{table*}[h]
    \centering
    \resizebox{0.78\textwidth}{!}{
    \begin{tabular}{l|c|c|ccc|c}
    \toprule
     Model & LLM params. & Frames & Finding & Summary & Reasoning & Overall \\
     \midrule
     \multicolumn{7}{c}{\textit{Baseline without visual information}} \\
     \midrule
     
    Baseline (GPT4o-text only~\cite{openai2024gpt4o}) & - & - &21.88 & 21.47 & 20.39 & 21.27 \\
    \midrule
    \multicolumn{7}{c}{\textit{Open-source MLLMs}} \\
     \midrule
    LongVA~\cite{zhang2024longva} & 7B & 32 & 49.41 & 40.38 & 37.19 & 42.33 \\
    Qwen2.5-VL~\cite{Qwen2VL} & 7B & 32 & 60.47 & 58.97 & 44.63 & 54.69 \\
    LLaVA-OneVision~\cite{li2024llavaonevision} & 7B & 32 & 67.76 & 52.56 & 44.90 & 55.08 \\
    MiniCPM-V2-6~\cite{yao2024minicpm} & 7B & 64 & 67.53 & 53.85 & 50.14 & 57.17 \\
    InternVideo2-chat~\cite{wang2024internvideo2} & 8B & 16 & 42.82 & 35.58 & 41.87 & 40.09 \\
    InternVL2~\cite{chen2024intervl2} & 8B & 32 & 53.88 & 44.55 & 46.56 & 48.33 \\
    %\hline
    LLaVA-NeXT-Video~\cite{zhang2024llavanextvideo} & 32B & 32 & 63.06 & 60.26 & 53.44 & 58.92 \\
    \midrule
    \multirow{4}{*}{LLaVA-Video~\cite{zhang2024llavanextvideo}}  & \multirow{4}{*}{72B} & 32 & 66.35 & 68.91 & 64.46 & 66.58 \\
    &  & 64 & 71.06 & 66.03 & 66.12 & 67.73 \\
     &  & 96 & 72.94 & 66.67 & 67.22 & 68.94 \\
     &  & 128 & 72.94 & 66.03 & 66.39 & 68.45 \\
    \hline 
    \multirow{4}{*}{Qwen2.5-VL~\cite{Qwen2VL}} & \multirow{4}{*}{72B} & 32 & 67.76 & 63.78 & 61.43 & 64.33 \\
    &  & 64 & 74.59 & 69.87 & 66.67 & 70.38 \\
     &  & 96 & 73.65 & 73.08 & 69.70 & 72.14 \\
     &  & 128 & 75.53 & 73.72 & 70.80 & 73.35 \\
    \midrule
    \multicolumn{7}{c}{\textit{Close-source MLLMs}} \\
    \midrule
    \multirow{3}{*}{GPT-4o~\cite{openai2024gpt4o}} & \multirow{3}{*}{-} & 32 & 70.59 & 66.35 & 59.78 & 65.57 \\
    & & 64 & 73.65 & 64.42 & 59.50 & 65.86 \\
    & & 128 & 75.06 & 63.14 & 62.26 & 66.82 \\
    \midrule 
    Gemini-1.5 Pro~\cite{Reid2024Gemini1U} & - & 1 fps & 75.29 & 67.31 & 66.39 & 69.66 \\
    Gemini-2.5 Pro~\cite{comanici2025gemini25} & - & - & 84.47 & 78.53 & 77.13 & 80.04 \\
    \midrule
     \multicolumn{7}{c}{\textit{Human Performance}} \\
    \midrule
    Human Performance & - & - & 93.41 & 94.55 & 95.04 & 94.27 \\
    \bottomrule
    \end{tabular}}
    \caption{Benchmark results for different MLLMs. We can observe that human performance substantially surpasses all SOTA MLLMs, while Gemini 2.5 Pro leads the best performance of 80.04\%. Some prediction cases are shown in the Appendix.}
    \label{tab:benchmark_results}
    \vspace{-6mm}
\end{table*}

\subsection{Experiment setting}
\textbf{MLLMs:}
We evaluate our collected dataset using open-source MLMMs, including InternVideo2~\cite{wang2024internvideo2}, LongVA~\cite{zhang2024longva}, InternVL2~\cite{chen2024intervl2}, %VILA-1.5~\cite{lin2023vila}, 
LLaVA-Next-Video~\cite{zhang2024llavanextvideo}, LLaVA-Video~\cite{zhang2024videoinstructiontuningsynthetic}, Qwen2.5-VL~\cite{bai2025qwen2}, which covers models from 7B to 72B parameters; and commercial models GPT-4o~\cite{openai2024gpt4o} , Gemini 1.5 Pro and 2.5 Pro~\cite{Reid2024Gemini1U,comanici2025gemini25}. 
We also compare the influence of the number of sampled frames on various models for those models that accept different numbers of frames as input.

\noindent\textbf{LLM performance:}
To ensure VideoAds serves as a reliable benchmark for video understanding, it is crucial to eliminate the possibility of MLLMs exploiting textual patterns rather than genuinely understanding visual content. Therefore, we conduct a baseline experiment using GPT-4o with text-only input, excluding all visual signals. Specifically, the model was prompted with instructions such as, 'Based only on the provided question, answer the following multiple-choice question,' without any indication that visual input was required. This process helps quantify the model's reliance on visual information versus textual bias.
% By establishing this text-only baseline, we quantify the extent of textual bias and ensure that VideoAds accurately evaluates temporal reasoning and multi-modal comprehension. 

\noindent\textbf{Human performance:}
To assess the performance of humans in such a challenging task, we recruit two master’s students as human experts and instruct them to complete the following tests: Given that most advertisement videos are designed to contain specific information for human, it is relatively easy to answer accurately by the evaluators. Each participant is instructed to watch each video once and then answer the corresponding VQA pairs without revisiting the video. This approach mimics the real-world constraints that MLLMs face, ensuring a fair comparison between human and model performance. 

% \begin{table}[h]
%     \centering
%     \resizebox{\linewidth}{!}{
%     \begin{tabular}{lcccc}
%     \toprule
%      &  Finding & Summary & Reasoning & Total \\
%     \hline
%      GPT4 & 21.88 & 21.47 & 20.39 & 21.27 \\
%     \hline
%      Human & 93.41 & 94.55 & 95.04 & 94.27 \\
%     \bottomrule
%     \end{tabular}}
    
%     \caption{Caption}
%     \label{tab:human_vs_LLM_performance}
% \end{table}

\noindent\textbf{Evaluation:}
To ensure a standardized and fair comparison across different MLLMs, we organize the VideoAds dataset into a four-option multiple-choice format. Each question is accompanied by one correct answer and three carefully curated distractors, which are semantically plausible but incorrect. %This four-option format also simplifies performance evaluation by allowing the use of prediction accuracy as a direct metric, making it easier to quantitatively compare models without ambiguity. 
We merge our dataset into the LMM-Eval package~\cite{zhang2024lmmsevalrealitycheckevaluation}, which has been widely used in MLLMs evaluation such that other researchers can use this dataset easily.

\subsection{Benchmarking results}

Table~\ref{tab:benchmark_results} presents a comprehensive comparison of model performance across three groups: LLM performance without visual input, MLLMs with video input, and human experts. This comparison provides several critical insights into the effectiveness of visual reasoning in video and highlights the significant gap between human cognitive abilities and current MLLMs.

%\subsubsection{LLM and Human Performance}
\textbf{LLM without visual information gives rise to low accuracy on VideoAds}: To assess the reliance on visual features, we conduct a baseline test using GPT-4o without any video input. The model achieves an accuracy of 21.27\%, which is close to the 25\% random guess baseline for a four-option multiple-choice format. This result confirms that VideoAds is effectively designed to eliminate trivial text-based shortcuts, ensuring that visual reasoning is indispensable for answering the questions. It also demonstrates that common-sense knowledge and language priors alone are insufficient to solve the complex visual and temporal reasoning tasks posed by this benchmark.

\textbf{Human performance substantially surpasses state-of-the-art MLLMs}: Human experts consistently outperform all tested MLLMs, achieving an impressive accuracy of 94.27\%. This highlights the inherent cognitive advantage of humans in narrative comprehension, causal inference, and high-level reasoning. Notably, humans perform best on the Visual Reasoning tasks, achieving 95.04\% accuracy, demonstrating their ability to synthesize information across disjointed scenes and understand implicit messaging strategies typical of advertisement videos. In stark contrast, most MLLMs struggle to reach even 70\% accuracy on the reasoning task, with the best-performing model, Qwen2.5-VL-72B, attaining 70.80\% accuracy, despite its large model size and state-of-the-art architecture. This substantial gap between human and model performance also highlights the key challenges raised by our VideoAds dataset.

\textbf{Current MLLMs excel in static visual recognition but struggle with complex reasoning, while humans give contradictory results}: From Table~\ref{tab:benchmark_results}, a clear performance disparity emerges across the three task types: visual finding, visual summary, and visual reasoning. Specifically, most MLLMs  perform the best on the visual finding tasks, followed by visual summary, with the worst performance observed on Visual Reasoning tasks. For example, GPT-4o achieves an accuracy of 75.06\% on the visual finding tasks but drops significantly to 62.26\% on the visual reasoning tasks with 128 frames. This pattern is consistent across other models,  such as LLaVA-Video and Qwen2.5-VL under different sampling frames. This performance gradient suggests that current MLLMs are more adept at static visual recognition and shallow temporal tracking but struggle with tasks requiring complex narrative reasoning. We also observe an interesting result for human performance in Table~\ref{tab:benchmark_results} that the performance on the most challenging reasoning tasks (accuracy 95.04\%) is even higher than the visual finding (accuracy 93.41\%). This is reasonable given that human experts find it easier to find the visual reasoning structures in advertisement videos at a high level but find it hard to remember every detail.

\textbf{State-of-the-art open source MLLMs can beat GPT-4o and Gemini 1.5 Pro, while Gemini 2.5 Pro continues to lead performance}: For a long time commercial LLMs~\cite{openai2023gpt4v,openai2024gpt4o,team2023gemini} can easily beat all opensource models in video understanding. VideoAds, however, produces the opposite result, with Qwen2.5-VL outperforming GPT-4o and Gemini 1.5 Pro in Table~\ref{tab:benchmark_results}. GPT-4o and Gemini 1.5 Pro can still achieve SOTA results (above 75\%) on the visual finding task but struggle with the visual summary and reasoning (around 66\%-67\%). However, Gemini 2.5 Pro, the most recent model, keep leading the performance with an accuracy of 80.04\%. This further emphasizes that rather than taking video as a grid of images, it is necessary to train MLLMs on video data specifically to extract the temporal relationship within the video like used in Gemini 2.5 Pro~\cite{comanici2025gemini25}. 

\textbf{Influence of increasing sampled frames varies across the models}:  When a video is simple, a limited set of sampled frames can include all necessary information. However, in complex videos like those in VideoAds, increasing the number of frames can significantly impact performance. We observe that Qwen2.5-VL benefits more from additional frames than GPT-4o and LLaVA-Video. The latter two models show notable improvements in visual finding tasks but only marginal gains in reasoning tasks when we increased the number of frames per video from 32 to 128. We hypothesize that,  due to high video complexity in VideoAds, it actually requires the MLLMs to process high fps and long context tokens, which are underexplored in the previous work~\cite{Qwen2VL,Reid2024Gemini1U,openai2024gpt4o,zhang2024llavanextvideo}.

\section{Discussion}
\label{sec:discussion}

% \subsection{Performance by Track}

% For example, Qwen2.5-VL-72B-Instruct achieves an accuracy of 67.76\% on the Visual Finding task but experiences a significant drop to 61.43\% on the Visual Reasoning task. This pattern is consistent across other models, such as LLaVA-Video-72B-Qwen2 and LLaVA-NeXT-Video-32B-Qwen, which exhibit a similar decrease in performance as the questions require higher-order reasoning and contextual understanding. In contrast, Visual Summary tasks maintain an intermediate level of accuracy, as they require event sequencing and thematic understanding without demanding multi-stage causal inference.  

% This performance gradient suggests that current MLLMs are more adept at static visual recognition and shallow temporal tracking but struggle with tasks necessitating complex narrative reasoning. The Visual Finding tasks primarily involve object identification, attribute recognition, and spatial awareness, which align well with the frame-based perception architecture of modern MLLMs. Conversely, Visual Reasoning tasks demand the model to synthesize information across non-contiguous scenes, infer implicit causal relationships, and understand nuanced emotional or thematic shifts.  The accuracy gap between Visual Finding and Visual Reasoning illustrates the cognitive challenge of advertisement videos, where high-level persuasion narratives and implicit messaging strategies are prevalent. Moreover, the results also highlight the effectiveness of VideoAds as a benchmark for evaluating temporal reasoning and narrative comprehension.

\subsection{Impact of Speech Audio Transcript}
To evaluate the influence of audio transcripts on model performance, we generate subtitles for each video using OpenAI Whisper~\cite{radford2023whisper}. We further test the performance of the models with transcripts and the results are shown in Table~\ref{tab:audio_influence}. Comparing transcript-enhanced models to those using only frame-based information, we observe that introducing audio transcripts significantly improves performance across all MLLMs. This suggests that many MLLMs for video are focusing on visual information, while the important cross-modality reasoning capability remains underexplored.

\begin{table}[!h]
    \centering
    \resizebox{\linewidth}{!}{
    \begin{tabular}{lcccccc}
    \toprule
    Model & Modality & Finding & Summary & Reasoning & Overall \\
    \midrule 
    \multirow{2}{*}{LongVA} & frames & 49.41 & 40.38 & 37.19 & 42.33 \\
    & + subtitle & 50.35 & 48.72 & 45.45 & 48.18 \\
    \midrule
    
    \multirow{2}{*}{LLaVA-Video} & frames & 66.35 & 68.91 & 64.46 & 66.58 \\
    & + subtitle & 71.53 & 71.47 & 72.18 & 71.73 \\
    \midrule
    \multirow{2}{*}{Qwen2.5-VL} & frames & 67.76 & 63.78 & 61.43 & 64.33 \\
     & + subtitle & 68.94 & 67.95 & 65.56 & 67.48 \\
    \midrule
    \multirow{2}{*}{GPT-4o} & frames  & 73.65 & 64.42 & 59.50 & 65.86 \\
      & + subtitle& 72.24 & 64.10 & 63.09 & 66.47 \\
    \bottomrule
    \end{tabular}}
    \caption{Impact of audio transcript: we can observe that a significant performance gain, particularly in the reason tasks can be achieved by various MLLMs with the help of audio transcript.}
    \label{tab:audio_influence}
    \vspace{-4mm}
\end{table}

\subsection{Impact of Chain of Thought}
The research on the influence of CoT on the multimodality data remains limited, and most studies focus on the image data~\cite{chen2023measuringcot,wei2022cot}. Here we present the first study of CoT's influence on video MLLMs using our challenging dataset. In our CoT prompting, we provided explicit intermediate questions to guide reasoning, rather than using generic 'think step-by-step' instructions. For example, for a visual reasoning task, we might pose intermediate questions like, 'Q1: What product appears in the first 5 seconds?' followed by 'Q2: What emotion does the character express at the end?' The model's answers to these intermediate questions were then used as context to generate the final response to the main question. We adopted this guided setup because many current MLLMs do not reliably support open-ended, multi-round conversational reasoning from a single generic prompt.  %We test the influence of CoT by introducing several reasoning steps for the challenging VQA tasks.
\begin{figure}[h]
    \centering
    \includegraphics[width=0.95\linewidth]{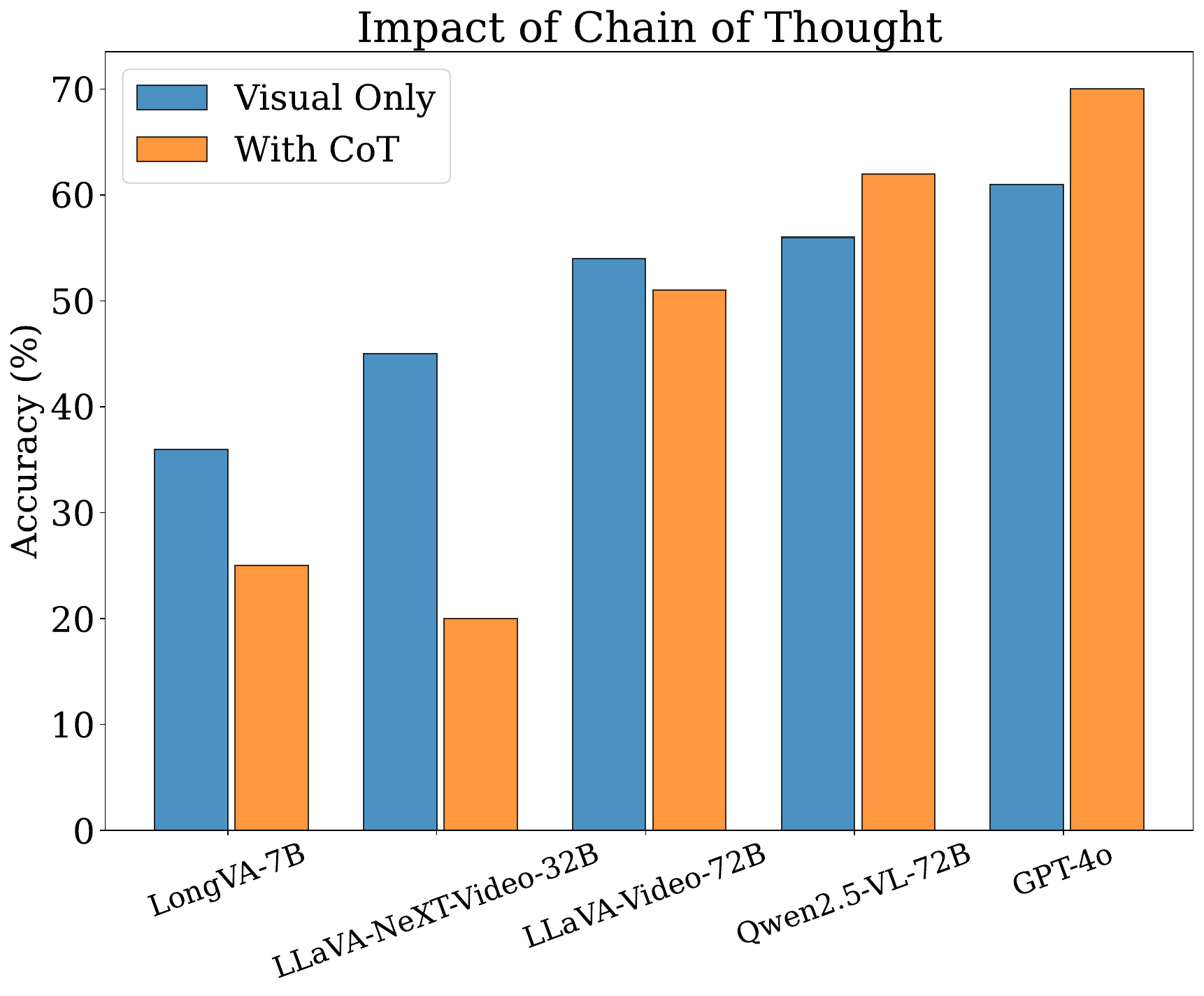}
    \caption{Impact of Chain of thought on the model's performance for challenging reasoning tasks, we can observe a variance in the model's performance along the model size.}
    \label{fig:enter-label}
    \vspace{-4mm}
\end{figure}

Interestingly, the influence of CoT differs across models: Generally, larger models tend to benefit more from the CoT as shown in the NLP field~\cite{ho2022largereason,wei2022chain}. We observe that LongVA-7B~\cite{zhang2024longva}, LLava-Next-32B~\cite{zhang2024llavanextvideo}, and LLava-Video-72B~\cite{zhang2024llavanextvideo} show a performance drop when dealing long reasoning contexts provided by CoT. However, models like Qwen2.5-VL~\cite{Qwen2VL} and GPT-4o~\cite{hurst2024gpt4o} show significant performance improvements. Particularly, GPT-4o increases its reasoning accuracy from 61\% to 70\%, indicating strong reasoning performance given CoT. This also underscores the importance of introducing long context and multi-round VQA in MLLM training~\cite{wei2022cot,chen2023measuringcot}.

\section{Conclusion}
\label{sec:conlusion}

In this work, we introduce VideoAds, the first benchmark dataset specifically designed to evaluate Multi-modality Large Language Models (MLLMs) on complex temporal reasoning and narrative understanding in advertisement videos. %Unlike traditional video benchmarks, VideoAds captures the intricate storytelling, rapid scene transitions, and implicit causal structures unique to advertisement narratives, providing an unprecedented challenge for current vision-language models. 
Our novel quantitative complexity metric provides a structured framework for evaluating temporal dynamics in video benchmarking. 
Extensive benchmarking experiments show that there is still a significant gap between human cognitive reasoning and the ability of current MLLMs to understand complex temporal structures like advertisement videos, highlighting the urgent need for advanced temporal modeling and narrative comprehension techniques.

\newpage
{
    \small
    \bibliographystyle{ieeenat_fullname}
    \bibliography{main}
}

%     \small
%     \bibliographystyle{ieeenat_fullname}
%     \bibliography{main}
% }
\end{document}